\documentclass[10pt,twocolumn,letterpaper]{article}

\usepackage{cvpr}
\usepackage{times}
\usepackage{epsfig}
\usepackage{graphicx}
\usepackage{amsmath}
\usepackage{amssymb}
\usepackage{algorithm}
\usepackage{graphicx}
\usepackage{algpseudocode}
\usepackage{bm}

\usepackage[breaklinks=true,bookmarks=false]{hyperref}

\cvprfinalcopy 

\def\cvprPaperID{****} 

\begin{document}

\title{Correlated and Individual Multi-Modal Deep Learning for
RGB-D Object Recognition}

\author{Ziyan Wang\\
Tsinghua University\\
{\tt\small zy-wang13@mails.tsinghua.edu.cn}
\and
Jiwen Lu\\
Tsinghua University\\
{\tt\small lujiwen@tsinghua.edu.cn}
\and 
Ruogu Lin\\
Tsinghua University\\
{\tt\small lrg14@mails.tsinghua.edu.cn}
\and 
Jianjiang Feng\\
Tsinghua University\\
{\tt\small jfeng@tsinghua.edu.cn}
\and
Jie Zhou\\
Tsinghua University\\
{\tt\small jzhou@tsinghua.edu.cn}
}

\maketitle

\begin{abstract}
   In this paper, we propose a correlated and individual multi-modal deep learning (CIMDL) method for RGB-D object recognition. Unlike most conventional RGB-D object recognition methods which extract features from the RGB and depth channels individually, our CIMDL jointly learns feature representations from raw RGB-D data with a pair of deep neural networks, so that the sharable and modal-specific information can be simultaneously and explicitly exploited. Specifically, we construct a pair of deep residual networks for the RGB and depth data, and concatenate them at the top layer of the network with a loss function which learns a new feature space where both the correlated part and the individual part of the RGB-D information are well modelled. The parameters of the whole networks are updated by using the back-propagation criterion. Experimental results on two widely used RGB-D object image benchmark datasets clearly show that our method outperforms most of the state-of-the-art methods.
\end{abstract}
\section{Introduction}
\begin{figure}[tb]
        \centering
        \includegraphics[width = 0.48\textwidth]{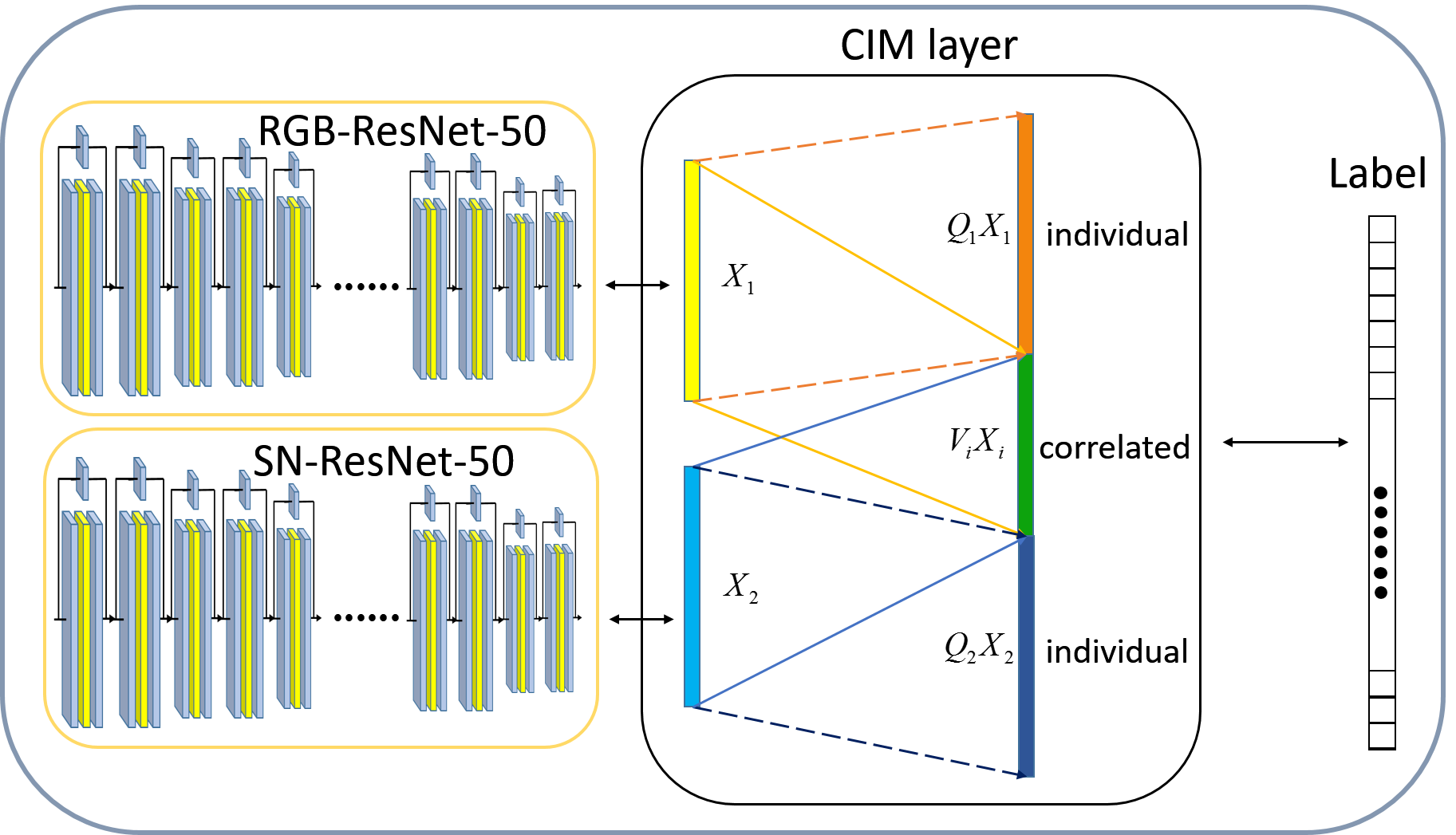}
        \caption{The pipeline of our proposed approach. We construct a two-way ResNet for RGB images and surface normal depth images for feature extraction. We design a multi-modal learning layer to learn the correlated part $V_iX_i$ and the individual parts $Q_iX_i$ of the RGB and depth features, respectively. We project the RGB-D features $X_1, X_2$ into a new feature space. The loss function enforces the affinity of the correlated part from different modalities and discriminative information of the modal-specific part, where the combination weights are also automatically learned within the same framework. (Best view in the color file)}
\end{figure}
Object recognition is one of the most challenging problems in computer vision, and is catalysed by the swift development of deep learning~\cite{DBLP:conf/cvpr/GirshickDDM14,DBLP:journals/corr/HeZRS15,NIPS2012_4824} in recent years. Various works have achieved exciting results on several RGB object recognition challenges~\cite{deng2009imagenet,Everingham10}. However, there are several limitations for object recognition using only RGB information in many real world applications, as it projects the 3-dimensional world into a 2-dimensional space which leads to inevitable data loss. To amend those shortcomings of RGB images, using depth images as a complimentary is a plausible way. The RGB image contains information of color, shape and texture while the depth contains information of shape and edge. Those basic features can serve both as a strength or weakness in object recognition. For example, we are able to tell the difference between an apple and a table simply by the shape information from depth. However it is ambiguous when it comes to figure out whether it is an apple or an orange just by depth. When an orange plastic ball and an orange are placed together, it is equally difficult for us to tell the difference just by RGB image. This means that a simple combination of features from two modalities sometimes jeopardizes the discriminability of feature. Therefore, we are supposed to choose those shared and specific features more wisely. Thus, we believe a more elaborated combination of modality-specific and modality-correlated features will generate a more discriminative representation.

There are two main procedures in RGB-D object recognition: feature representation~\cite{Abdel-Hakim:2006:CSD:1153171.1153685,Bay:2008:SRF:1370312.1370556,Leung:2001:RRV:543015.543017,Lowe:2004:DIF:993451.996342} and object matching~\cite{DBLP:journals/ml/CortesV95,DBLP:conf/nips/ElfadelWW93,DBLP:conf/icdar/Ho95}. Compared with object matching, feature representation affects the performance of the object recognition system significantly, because real-world objects usually suffer from large intra-class discrepancy and inter-class affinity. A variety of methods have been proposed for RGB-D object representation recently, and they can be mainly classified into two categories: hand-crafted methods and learning-based methods. Methods in the first category design an elaborated hand-crafted descriptor for both the RGB and depth channels~\cite{DBLP:conf/iros/BoRF11, conf/iccvw/BrowatzkiFGBW11, lai_icra11a} for feature extraction. However, these hand-crafted methods usually require large amount of domain-specific knowledge, which is inconvenient to generalize to different datasets. Methods in the second category employ some machine learning techniques to learn feature representations in a data-driven manner, so that more data-adaptive discriminative information can be exploited~\cite{blumicra2012,Bo12unsupervisedfeature,lai_icra11b,SocherEtAl2012:CRNN,Wang_2015_ICCV}. However, most existing learning-based methods consider the RGB and depth information from two different channels individually, which ignores the sharable property and the interaction relationship between these two modalities. To address this, multi-modal learning approach recently has been presented for RGB-D object recognition. 

In this paper, we propose a correlated and individual multi-modal deep learning (CIMDL) method for RGB-D object recognition. Specifically, we develop a multi-modal learning framework to learn discriminative features from both the correlated and individual parts, and automatically learn the weights for different feature components in a data-driven manner. The basic pipeline of our proposed CIMDL is illustrated in Figure~1. This layer is designed for three purposes: 1) generating the correlated part between these two modalities, 2) extracting the discriminative part of features from both of these two modalities and 3) learning the weights for the correlated and individual parts automatically for feature combination. The parameters of the whole network are updated by using the back-propagation criterion. Experimental results on the RGB-D object~\cite{lai_icra11a} and 2D3D~\cite{conf/iccvw/BrowatzkiFGBW11} datasets are presented to show the effectiveness of the proposed approach.
\section{Related Work}
There are many visual tasks based on the RGB-D input~\cite{Cheng_2015_ICCV,Feng_2016_CVPR,DBLP:conf/eccv/HuZMWL16,Rogez_2015_ICCV,DBLP:journals/pami/WangLWY14,DBLP:journals/corr/WangWTSW16,Zhu_2016_CVPR}. Most conventional RGB-D object recognition methods use hand-crafted feature descriptors for object matching. For example, Lai \textit{et al.}~\cite{lai_icra11a} exploited a bunch of hand-crafted features including color histograms~\cite{Abdel-Hakim:2006:CSD:1153171.1153685}, textons~\cite{Leung:2001:RRV:543015.543017} and SIFT~\cite{Lowe:2004:DIF:993451.996342} for the RGB channel, spin images~\cite{765655} and SIFT~\cite{Lowe:2004:DIF:993451.996342}, and multiple shape features for the depth channel for feature representation. Finally, they concatenated these features together as the final representation for recognition. Bo \textit{et al.} proposed a kernel descriptor method~\cite{DBLP:conf/iros/BoRF11} which combines several RGB-D features such as 3D shape, depth edge and texture for RGB-D object recognition.

In recent years, learning-based features have aroused more and more attention in RGB-D object recognition. For example, Blum~\textit{et al.}~\cite{blumicra2012} used an unsupervised learning method to obtain a codebook for feature encoding. Bo~\textit{et al.}~\cite{bo_nips11} presented a Hierarchal Matching Pursuit (HMP) method to learn advanced representations from local patches. Socher~\textit{et al.}~\cite{SocherEtAl2012:CRNN} presented a cascaded network by integrating CNN and RNN to learn a concatenated feature for RGB and depth image. Browazki \textit{et al.}~\cite{conf/iccvw/BrowatzkiFGBW11} fused several SVMs-trained features by utilizing a multi-layer perception. Lai~\textit{et al.}~\cite{lai_icra11b} devised a distance metric learning approach~\cite{malisiewicz-cvpr08,NIPS2003_2366} to fuse heterogeneous feature representations for RGB-D object recognition. However, most of them learn features from the color and depth channels individually, and construct a final representation by a simple concatenation, which ignores the physical meanings of different feature modalities and their potential relationship.

Deep learning has also been employed for RGB-D visual analysis or in  recent years. For example, Gupta~\textit{et al.}~\cite{DBLP:conf/eccv/GuptaGAM14} proposed an approach to encode the depth data into three channels: horizontal disparity, height above ground, angle between point normal and inferred gravity. Then, they trained CNNs on these three-channel instead of the ordinal depth images for RGB-D object recognition and segmentation. Couprie \textit {et al.}~\cite{DBLP:journals/corr/abs-1301-3572} presented a multi-scale CNN for RGB-D scene labeling based on a hierarchical feature method. Wang \textit{et al.}~\cite{wang2015designing} designed a deep neural network for surface normal prediction. However, these methods ignored the relationship between data from different modalities because the RGB and depth information are only simply concatenated. 

More recently, several multi-modal deep learning methods have been proposed to make better use of information from different modalities for various visual analysis tasks. For example, Srivastava~\textit{et al.}~\cite{DBLP:conf/nips/SrivastavaS12} proposed a multi-modal Deep Boltzmann Machine approach, where a concatenated layer was added to connect DBMs from different modalities to learn multi-modal feature representations jointly. Eitel~\textit{et al.}~\cite{DBLP:conf/iros/EitelSSRB15} proposed a two-stream CNN model combined with a fusion layer for RGB-D object recognition. Wang~\textit{et al.}~\cite{Wang_2015_ICCV} proposed a multi-modal feature deep learning approach by exploiting the shareable properties of RGB and depth images for RGB-D object recognition. Inspired by Wang's work, Zhu~\textit{et al.}~\cite{Zhu_2016_CVPR} proposed a discriminative multi-modal feature learning method for RGB-D scene recognition. Lenz \textit{et al.}~\cite{DBLP:journals/ijrr/LenzLS15} presented a multi-modal deep learning approach for robotic grasp detection, where the stacked auto-encoders were used for multi-modal feature learning. Jou \textit{et al.}~\cite{DBLP:journals/corr/JouC16} proposed a cross-residual learning for multitask learning.

\section{Proposed Approach}
\subsection{Baseline Architecture}
Several ways of CNN-based methods have been proposed for RGB-D object recognition. For example, Gupta \textit{et al.}~\cite{DBLP:conf/eccv/GuptaGAM14} extracted features from RGB and depth images independently and concatenated them as the final features for object detection, where both the RGB CNN and the depth CNN are fine-tuned on the model which was pre-trained on the ImageNet dataset~\cite{deng2009imagenet}. Another method is to fuse the second fully connected layer of the RGB network and the depth network so that the supervised information can be back-propagated for both modalities. As the residual network has proved its strength over conventional CNN architecture~\cite{DBLP:journals/corr/HeZRS15}, we adopt the ResNet as baseline architecture for both RGB and depth channel and train the networks independentlyto extract features from the last pooling layer of the ResNet. For the depth network, we adopt the surface normals instead of the depth image for network training, which can be fine-tuned on the RGB pre-trained model.
\begin{figure}[tb]
        \centering
        \includegraphics[width = 0.48\textwidth]{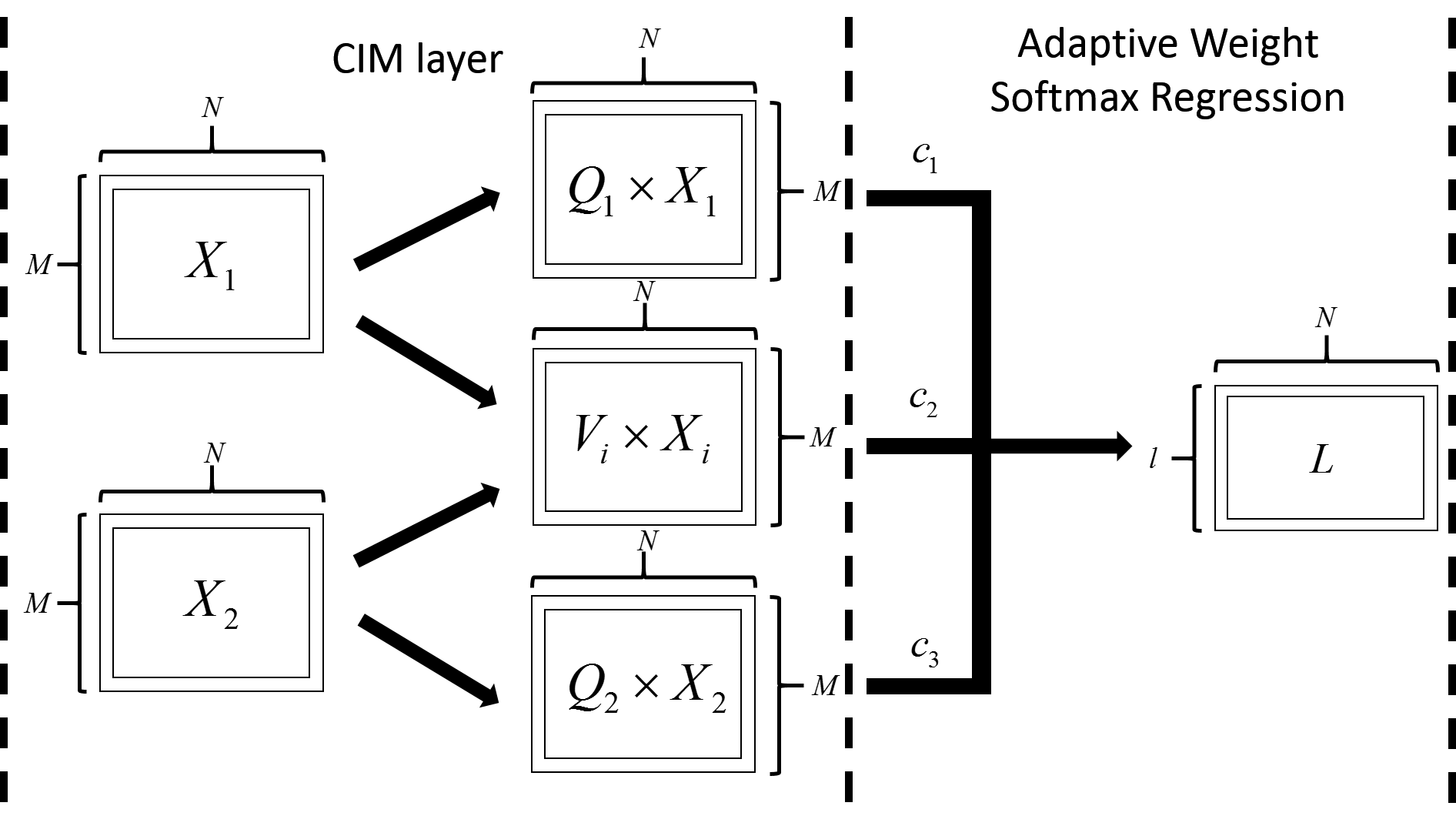}
        \caption{The architecture of the last layer of our network, where $V_i$ and $Q_i$ denote the mapping matrices that map the original feature into a correlated and individual feature space. The output of the last layer is concatenated into a new feature vector which is assigned with different weights $c_i$ for different components in the learning framework. }
\end{figure}
\subsection{Multi-modal Deep Learning Model}
We develop a multi-modal deep architecture for RGB-D feature learning, which consists of two residual networks. Specifically, we adopt two models which are pre-trained on the ImageNet dataset~\cite{deng2009imagenet} and finetune them on our RGB training dataset and the depth training dataset to generate the parameters of RGB-ResNet and SN-ResNet layers, respectively. Then, we feed the outputs of the last pooling layer from both RGB-ResNet and SN-ResNet into our correlated and individual multi-modal learning structure. In our new structure, we replace the original softmax layer with our new CIMDL layer which will be detailed later. The pipeline of our proposed method is shown in Figure~1. Instead of directly putting the depth image to the ResNet network, we also extract the surface normal of the depth information, which encodes it as a three-channels representation. We empirically find that the surface normal results in better recognition performance than the raw depth data in feature learning.

Figure~2 shows the architecture of the last layer of our network, where $X_i = [x_{i1} , x_{i2}, \cdots x_{iN}] {\in} \mathbf{R}^{M\times N}$ denote the activations of the second fully connected layer of RGB-ResNet and SN-ResNet with $N$ images in one data batch, $Q_i{\in} \mathbf{R}^{M\times M}$ and $V_i {\in} \mathbf{R}^{M\times M}$ are mapping matrices which transfer original features into the modal-specific domain and the correlated domain, and $L \in \mathbf{R}^{l\times N}$ denotes the label matrix with $l$ classes.

The physical meaning of our multi-modal learning model is to leverage the correlated properties of the RGB and depth information, enforce the modal-specific property of both modalities and adjust the weights for different parts of the feature in recognition. Therefore, the final purpose of our proposed method is to learn two mapping matrices $V$ and $Q$ to map the original feature into the correlated feature space and the individual feature space. Hence, there are three key characteristics in our model: 1) a multi-modal deep learning strategy which automatically decomposes features into a correlated part and an individual part, 2) ensuring the discriminative power and orthogonality of the correlated part and the individual part and 3) learning the weights for different parts in a data-driven manner to improve the recognition performance.
\subsection{Objective Function}
We first learn the mapping matrices $V_{i}(i = 1,2)$ for both modalities to map the original features into the correlated feature space, where we expect to minimize the difference between the correlated parts from two datasets, respectively. Let $X_i {\in} \mathbf{R}^{M\times{N}}$ be the $M$-dimensional activations of the last pooling layer in one batch with $N$ images, where $i = 1, 2$ corresponds to the RGB channel and depth channel, respectively. Our goal is to learn discriminative feature representations to achieve two objectives: 1) some information are shared by different modalities, and 2) some modal-specific information are exploited for each modality individually. To achieve this, we formulate the following objective function:
\begin{align}
\begin{split}
 & \mathop { \min }\limits_{{{\boldsymbol{V}}_1},{{\boldsymbol{V}}_2}} {{\left\| {{{\boldsymbol{V}}_1}{{\boldsymbol{X}}_1}-{{\boldsymbol{V}}_2}{{\boldsymbol{X}}_2} } \right\|}_F^2}
\end{split}
\end{align}
where $ {\left\| \cdot \right\|}_{F} $ denotes the Frobenius norm. By minimizing the above objective function, the mapping matrix increases the similarity of correlated part from both modalities. Besides the correlated part, the modal-specific feature is also an essential part of the feature $X_i(i = 1, 2)$. Hence, we present the following criterion to achieve this good:
\begin{align}
\begin{split}
{{\boldsymbol{X}}_i} = {{\boldsymbol{C}}_i} +{{\boldsymbol{A}}_i}   (i=1,2)
\end{split}
\end{align}
where $C_i$ is the correlated part and ${{A}_i}$ denotes the modal-specific components of the \textit{i}th modality's feature. The correlated part is the component of the original feature $X_i$. Both the correlated feature and model-specific feature reconstruct the original feature as follows:
\begin{align}
{{\boldsymbol{C}}_i}={{\boldsymbol{V}}_i^T}{{\boldsymbol{V}}_i}{{\boldsymbol{X}}_i}  (i=1,2)
\\ {{\boldsymbol{A}}_i}={{\boldsymbol{Q}}_i^T}{{\boldsymbol{Q}}_i}{{\boldsymbol{X}}_i}  (i=1,2)
\end{align}
where $V_i {\in} \mathbf{R}^{M\times{M}}$ is the mapping matrix to project the original feature into the correlated domain and ${{Q_i}{X}_i}$ denotes the individual component of the \textit{i}th modality's feature. Since these two matrices ${V_i}$ and ${Q_i}$ map the original feature into the correlated domain and the modal-specific domain, respectively, they should be unrelated and not contaminated by each other. Therefore, the mapping matrices contain bases that come from discrepant space and should be orthogonal to each other. Therefore, we present the following constraints:
\begin{align}
\begin{split}
{{\boldsymbol{V}}_i^T{{\boldsymbol{Q}}_i}=0,(i=1,2)}
\end{split}
\end{align}
Consequently, we formulate the objective function of our multi-modal feature learning as two parts. The first part enforces the features of two domains to share a congruity part after the mapping $V$. The second part is the softmax loss of our network. The first constraint ensures the reconstruction of the original feature, and the second constraint engenders discrepancy between the correlated part and the modal-specific part, described as below:
\begin{align}
\begin{split}
  \mathop { \min }\limits_{{{\boldsymbol{V}}_1},{{\boldsymbol{V}}_2},{{\boldsymbol{Q}}_1},{{\boldsymbol{Q}}_2}} & {\boldsymbol{J}={{\left\| {{{\boldsymbol{V}}_1}{{\boldsymbol{X}}_1}-{{\boldsymbol{V}}_2}{{\boldsymbol{X}}_2} } \right\|}_F^2}} \\
 + & { \mu\boldsymbol{\mathcal{L}(softmax(W,T,L,c))}}, \\
  s.t. \quad &  {{\boldsymbol{X}}_i} = {{\boldsymbol{V}}_1^T}{{\boldsymbol{V}}_i}{{\boldsymbol{X}}_i} + {{\boldsymbol{Q}}_1^T}{{\boldsymbol{Q}}_i}{{\boldsymbol{X}}_i}, \\
&  {{{\boldsymbol{V}}_i^T}{{\boldsymbol{Q}}_i}=0,(i=1,2)}
\end{split}
\end{align}
where $W=[W_1,W_2,W_3],W_i\in \mathbf{R}^{l\times M}$ are the weights of the last softmax layer for different modalities, $T$ is the activation of the CIMDL layer. In Figure~2, the matrices in the middle show the detail of $T$. $c$ is weight for different parts of the feature $T$ for softmax regression which will be introduced later. Here we incorporate supervised learning by minimizing the loss of softmax regression from $\boldsymbol{WT}$ to $L$.
\subsection{Optimization}
In this work, we adopt an alternating optimization for all those variables $V_i,Q_i,W$ and $c$. Following the Lagrange multiplier and gradient decent criterion, we obtain a local optimal solution for (6). Based on (6), we construct a Lagrange function as follows:
\begin{align}
\begin{split}
& \mathcal{L}(\alpha,\sigma,\delta,\theta,\mu)={{\left\| {{{\boldsymbol{V}}_1}{{\boldsymbol{X}}_1}-{{\boldsymbol{V}}_2}{{\boldsymbol{X}}_2} } \right\|}_F^2}\\
& +\sum_{i=1,2} {\alpha_i} {{\left\| {{\boldsymbol{X}}_i} - { {{\boldsymbol{V}}_i^T}{{\boldsymbol{V}}_i}{{\boldsymbol{X}}_i}-{{\boldsymbol{Q}}_i^T}{{\boldsymbol{Q}}_i}{{\boldsymbol{X}}_i} } \right\|}_F^2}\\
& + \sum_{i=1,2} {\sigma_i} {{\left\| {{{\boldsymbol{V}}_i^T}{{\boldsymbol{Q}}_i}} \right\|}_F^2} \\
& + \sum_{i=1,2}{ \delta_i  g({{\boldsymbol{V}}_i}{{\boldsymbol{X}}_i})  + \theta_i  \boldsymbol{g}({{\boldsymbol{Q}}_i}{{\boldsymbol{X}}_i})} \\
& + \mu \boldsymbol{h}({\boldsymbol{W}}{\boldsymbol{T}} - {\boldsymbol{L}}) + \eta{\left\| {\boldsymbol{W}} \right\|}_{2,1}
\end{split}
\end{align}
where $ {\left\| \cdot \right\|}_{F} $ denotes the Frobenius norm. $\alpha_i$ and $\sigma_i$ are positive Lagrange multipliers associated with the linear constrains ${{\boldsymbol{X}}_i} = {{\boldsymbol{V}}_i^T}{{\boldsymbol{V}}_i}{{\boldsymbol{X}}_i} + {{\boldsymbol{Q}}_i^T}{{\boldsymbol{Q}}_i}{{\boldsymbol{X}}_i}$ and ${{{\boldsymbol{V}}_i^T}{{\boldsymbol{Q}}_i}=0}$. The first term of the objective function intends to minimize the difference of the correlated part that generated from the color modality and the depth modality separately. The second term regularizes the feature's ability to be reconstructed by its correlated part and modal-specific part. The third part ensures the mutual orthogonality by regulating the inner product of the two transfer matrices ${{\boldsymbol{V}}_i}$ and ${{\boldsymbol{Q}}_i}$. The last part of the objective function are regularizations of ${{\boldsymbol{V}}_i}{{\boldsymbol{X}}_i}$ and ${{\boldsymbol{Q}}_i}{{\boldsymbol{X}}_i}$ where $g(\cdot)=log(cosh(\cdot))$ denotes the smooth $L_1$ penalty function \cite{DBLP:conf/nips/LeKNN11}.

By applying the gradient decent algorithm, we update the variables $V_i$, $Q_i$ and $W$ with the same learning rate. First, we update $V_i$,$Q_i$. The derivative of the Lagrange function $\mathcal{L}(\alpha,\sigma,\delta,\theta,\mu)$ with respect to $V_1$ can be expressed as
\begin{align}
\begin{split}
& \frac{\partial \boldsymbol{\mathcal{L}}}{\partial \boldsymbol{V}_1}  = 2[ ({\boldsymbol{V}_1}{\boldsymbol{X}_1} - {\boldsymbol{V}_2}{\boldsymbol{X}_2}){\boldsymbol{X}_1^T} + \sigma_1{\boldsymbol{Q}_1}{\boldsymbol{Q}_1^T}{\boldsymbol{V}_1} \\
& - 2\alpha_1 {\boldsymbol{V}_1}{\boldsymbol{X}_1}({\boldsymbol{X}_1} - {\boldsymbol{V}_1^T}{\boldsymbol{V}_1}{\boldsymbol{X}_1} - {\boldsymbol{Q}_1^T}{\boldsymbol{Q}_1}{\boldsymbol{X}_1})^T \\
& + \delta_1 {\boldsymbol{g}'}({\boldsymbol{V}_1}{\boldsymbol{X}_1}) \\
& + 0.5\mu {\boldsymbol{c}_1}{\boldsymbol{W}_1^T}({\boldsymbol{L}}-{\boldsymbol{P}}){\boldsymbol{X}_1^T}]
\end{split}
\end{align}
where $g'(\cdot)$ is the derived function of smooth $L_1$ penalty function. $c_1$ is the weight of the correlated part in the regression.
\begin{align}
\begin{split}
P= & exp(-[{\boldsymbol{c}_1}{\boldsymbol{W}_1},{\boldsymbol{c}_2}{\boldsymbol{W}_2},{\boldsymbol{c}_3}{\boldsymbol{W}_3}]  \\ & \times[\frac{{\boldsymbol{V}_1}{\boldsymbol{X}_1}+{\boldsymbol{V}_2}{\boldsymbol{X}_2}}{2},{\boldsymbol{Q}_1}{\boldsymbol{X}_1},{\boldsymbol{Q}_2}{\boldsymbol{X}_2}])
\end{split}
\end{align}

According to the gradient descent rule, ${\boldsymbol{V}_1}$ is updated as
\begin{align}
\begin{split}
{\boldsymbol{V}_1}[t+1]={\boldsymbol{V}_1}[t]-\gamma\frac{\partial \boldsymbol{\mathcal{\mathcal{L}}}}{\partial {\boldsymbol{V}_1}[t]}
\end{split}
\end{align}

%

The derivative of the Lagrange function  $\mathcal{L}(\alpha,\sigma,\delta,\theta,\mu)$ with respect to $Q_1$ can be expressed as
\begin{align}
\begin{split}
& \frac{\partial \boldsymbol{\mathcal{L}}}{\partial \boldsymbol{Q}_1} = 2[ - \alpha_1 {\boldsymbol{Q}_1}{\boldsymbol{X}_1}({\boldsymbol{X}_1} - {\boldsymbol{V}_1^T}{\boldsymbol{V}_1}{\boldsymbol{X}_1} -\\ & {\boldsymbol{Q}_1^T}{\boldsymbol{Q}_1}{\boldsymbol{X}_1})^T 
+ \sigma_1{ \boldsymbol{V}_1}{\boldsymbol{V}_1^T}{\boldsymbol{Q}_1}
 + \theta_1 \boldsymbol{g}'({\boldsymbol{Q}_1}{\boldsymbol{X}_1}) \\ & + \mu {\boldsymbol{c}_2}{\boldsymbol{W}_2}^T({\boldsymbol{L}}-{\boldsymbol{P}})*{\boldsymbol{X}_1^T}]
\end{split}
\end{align}

According to the gradient descent rule, ${\boldsymbol{Q}_1}$ is updated as
\begin{align}
\begin{split}
{\boldsymbol{Q}_1}[t+1]={\boldsymbol{Q}_1}[t]-\gamma\frac{\partial \boldsymbol{\mathcal{L}}}{\partial {\boldsymbol{Q}_1}[t]}
\end{split}
\end{align}

%

As ${\boldsymbol{V}_2}$ and ${\boldsymbol{Q}_2}$ share similar forms with ${\boldsymbol{V}_1}$ and ${\boldsymbol{Q}_1}$, we just omit how we update those variables. 

Having completed updating the mapping matrices $V_i$ and $Q_i$, we keep them fixed and update the regression matrix $W$. According to the work of \cite{DBLP:conf/nips/NieHCD10}, the derivative of the Lagrange function $\mathcal{L}(\alpha,\sigma,\delta,\theta,\mu)$ with respect to $W$ is
\begin{align}
\begin{split}
 \frac{\partial \boldsymbol{\mathcal{L}}}{\partial \boldsymbol{W}} = 2(\mu ({\boldsymbol{L}}-{\boldsymbol{P}}){\boldsymbol{T}^T} + \eta {\boldsymbol{E}}{\boldsymbol{W}})
\end{split}
\end{align}
where $\boldsymbol{E}$ is a diagonal matrix with $e_{jj}=0.5{\left\|w_j\right\|_2}$, $W=[w_1,w_2,\cdots,w_M]$ and $w_i {\in} \mathbf{R}^{l\times{1}}, i = \{1,2,\cdots,M\}$. According to the gradient descent criterion, we update $W$ by
\begin{align}
\begin{split}
{\boldsymbol{W}}[t+1]={\boldsymbol{W}}[t]-\gamma\frac{\partial \boldsymbol{\mathcal{L}}}{\partial {\boldsymbol{W}}[t]}
\end{split}
\end{align}

The correlated part of two distinct dataset feature plays a significant part in extracting the shareable information but not equally powerful as the modal-specific part. Therefore, we design adaptive weights for different part of the fused feature, where $c_1,c_2$ and $c_3$ correspond to the correlated parts, the modal-specific part for RGB and the modal-specific part for surface normal respectively. With the feature representation of $W,V_i$ and $Q_i$ updated and fixed, we update the adaptive parameter $c$ according to the following rules:
\begin{align}
& {\boldsymbol{c_1}={\left\| exp(-\boldsymbol{W_1} * (\boldsymbol{V_1}\boldsymbol{X_1}+\boldsymbol{V_2}\boldsymbol{X_2})) - L \right\|}_F^p} \\
& {\boldsymbol{c_2}={\left\| exp(-\boldsymbol{W_2} * \boldsymbol{Q_1}\boldsymbol{X_1}) - L \right\|}_F^p} \\
& {\boldsymbol{c_3}={\left\| exp(-\boldsymbol{W_3} * \boldsymbol{Q_2}\boldsymbol{X_2}) - L \right\|}_F^p}
\end{align}

We regularize the vector $c=[c_1,c_2,c_3]$ and make sure that $c_1 + c_2 + c_3 = 1$. Iteratively, we automatically select the weights for different parts of our learned feature. Our proposed CIMDL method is summarized in \textbf{Algorithm 1}.

\renewcommand{\algorithmicrequire}{\textbf{Input:}}
\renewcommand{\algorithmicensure}{\textbf{Output:}}
\begin{algorithm}[tb]
\caption{: CIMDL} \label{alg}
\begin{algorithmic}[1]
\Require
Training set from both modality$\mathbf{X_1},\mathbf{X_2}$,ground truth label matrix $\mathbf{L}$
\Ensure
Mapping matrix $\mathbf{V_i},\mathbf{Q_i}(i=1,2)$,projection matrix $\mathbf{W}=[\mathbf{W_1},\mathbf{W_2},\mathbf{W_3}]$, adaptive weights $\mathbf{c}=[\mathbf{c_1},\mathbf{c_2},\mathbf{c_3}]$
\State Initialize $\mathbf{W},\mathbf{V_1},\mathbf{V_2},\mathbf{Q_1},\mathbf{Q_2},\mathbf{c}$
\While{not converge}
\State Fix $\mathbf{W},\mathbf{c},$ \\
\quad{\quad{}} Update $\mathbf{V_1},\mathbf{V_2}$ according to (10).
\State Fix $\mathbf{W},\mathbf{c},$ \\
\quad{\quad{}} Update $\mathbf{Q_1},\mathbf{Q_2}$ according to (12).
\State Fix $\mathbf{V_1,V_2},\mathbf{Q_1,Q_2},\mathbf{c},$ \\
\quad{\quad{}} Update $\mathbf{W}$ according to (14)
\State Fix $\mathbf{V_1,V_2},\mathbf{Q_1,Q_2},\mathbf{W},$ \\
\quad{\quad{}} Update $\mathbf{c}$ according to (15)-(17).
\EndWhile \\
\Return
$\mathbf{W},\mathbf{V_1,V_2},\mathbf{Q_1,Q_2},\mathbf{c}$
\end{algorithmic}
\end{algorithm}

\subsection{Discussion}

In this section, we discuss the relationship and difference between our approach and several existing related methods:

\textbf{Differences from other deep learning methods~\cite{DBLP:journals/corr/abs-1301-3572,DBLP:conf/eccv/GuptaGAM14,DBLP:conf/icra/SchwarzSB15}:}. In~\cite{DBLP:journals/corr/abs-1301-3572}, the depth information is used as an extra channel which are concatenated with RGB image at the input. Both~\cite{DBLP:conf/icra/SchwarzSB15} and~\cite{DBLP:conf/eccv/GuptaGAM14} employed multi-stream architecture for their RGB channel and depth channel while they ignore the relationship between the RGB and depth information. Those works treat RGB and depth images in-differentially. The main difference between our proposed method and those deep learning based methods is that we explore the relationship between RGB and depth data and utilize both the correlated and individual information wisely.

\textbf{Difference from other multi-modal learning methods~\cite{DBLP:conf/iros/EitelSSRB15,SocherEtAl2012:CRNN,DBLP:conf/nips/SrivastavaS12,Wang_2015_ICCV}:}  CNN-RNN~\cite{SocherEtAl2012:CRNN} is a cascade architecture containing both CNN and RNN. Multi-modal DBM~\cite{DBLP:conf/nips/SrivastavaS12} trained two specific deep boltzmann machine for the text and image modalities. Comparing with these methods, our method use ResNet for feature learning of different modalities. Fus-CNN~\cite{DBLP:conf/iros/EitelSSRB15} employed a two stream CNN structure and adopted a three-step training method which includes both joint training and individual training. However, comparing with our method, they failed to leverage the information that is shared by both modalities. MMSS~\cite{Wang_2015_ICCV} proposed a multi-modal deep learning method which can utilize both sharable and specific information of different modalities. However, they focus on learning the shareable feature and manually set the ratio for shareable and specific part. In our work, we enforce the learning of modal-specific part by exerting constraints over those two parts and learn how to utilize both correlated and individual feature in a data-driven manner.

\section{Experiments}

We conducted experiments on two datasets including the RGB-D Object Dataset~\cite{lai_icra11a} and the 2D3D Dataset~\cite{conf/iccvw/BrowatzkiFGBW11} for RGB-D object recognition, to evaluate the effectiveness of our proposed method. The followings describe the detailed experimental settings and results.

\subsection{Datasets}

\textbf{RGB-D Object Dataset:} 
The RGB-D dataset is a large dataset containing 51 different classes of 300 distinct objects shoot from multiple views.There are 207,920 RGB-D image frames, with roughly 600 images per object. In our experiments, we conducted a down-sampling from every 5 consecutive frames of the video. We run the 10 random splits provided by~\cite{lai_icra11a}, where each of those splits covers the whole 51 classes with different objects. Finally, it came out with approximately 51 different test objects. We conducted experiments on those different splits, where there are 34000 images in average for training and 6900 images in average for testing.

\textbf{2D3D Dataset:} The 2D3D dataset includes 154 objects in 14 different classes.Following the same settings in~\cite{conf/iccvw/BrowatzkiFGBW11}, we divided the dataset into the training part and the testing part, containing 6 objects of each class. Finally, 1476 RGB-D images from 82 objects are used for training and 1332 RGB-D images from 74 objects are employed for testing.

\subsection{Implementat Details}

\textbf{Architecture of ResNets:} Our experiments were performed on the Caffe framework~\cite{jia2014caffe}. We adopted a 50-layers ResNet structure described in~\cite{DBLP:journals/corr/HeZRS15}. The two modalities shared the same network architecture before the last pooling layer. For both the RGB modality and the surface normal modality, the input images were resized to $224\times{224}\times{3}$.

\textbf{Finetuning:} The ResNet model~\cite{DBLP:journals/corr/HeZRS15} was pre-trained on the ImageNet dataset~\cite{deng2009imagenet} both for RGB and SN channels. Our finetuning setting is listed as follows. The learning rate was initialized at $0.001$ and decreased by a factor of 0.1 every $5,000$ iterations. We finetuned the model with $15,000$ iterations with a batch size of $16$. The whole finetuning part was done by a Tesla K40c and it took nearly $3$ hours for the whole finetuning procedure.

\textbf{Parameters Setting:} In the Lagrange function of our multi-modal learning model, $\alpha$ and $\sigma$ are the reconstruction and orthogonal controlling parameters, and they were set as follows: $ \{ \alpha_i = \frac{0.5}{N}| i=1,2 \}$ and $ \{ \sigma_i=\frac{0.5}{M}|i=1,2 \}$, where $N$ denotes the size of training samples and $M$ denotes the size of the feature vectors, $\mu$ is the softmax regression controlling parameter and it is set as $\frac{10}{N}$. $\delta,\theta,\eta$ are regularization parameters and they were empirically set as $ \{ \delta_i=\frac{0.005}{N}|i=1,2 \}$, $\{ \theta_i=\frac{0.005}{N}|i=1,2\}$ and $\eta = 1$, the learning rate for $W$ is set empirically as $10^{-3}$ and for $V_i,Q_i$ as $10^{-5}$, respectively. All parameters were set the same for experiments both on RGB-D object dataset and 2D3D dataset.

\subsection{Results on RGB-D Object Dataset}

\begin{figure*}[tb]
        \centering
        \includegraphics[width = 0.9\textwidth]{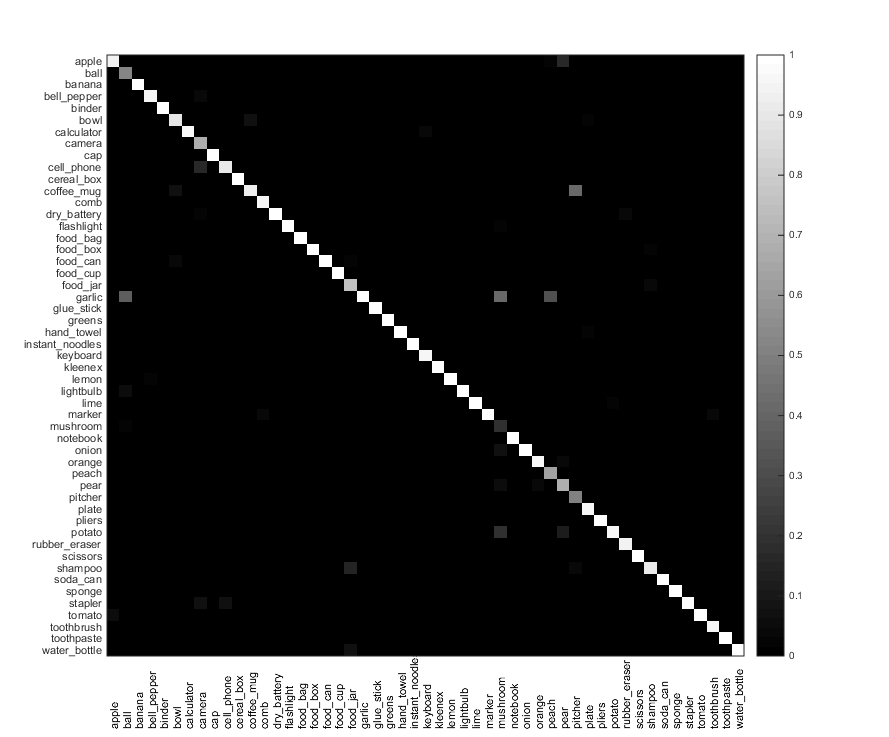}
        \caption{The confusion matrix of our method on the RGB-D object dataset, where the vertical axis indicates the ground truth label and the horizontal axis indicates the predicted label, respectively.}
\end{figure*}


\textbf{Comparison with Deep Learning Baselines:}
We first constructed several deep learning baselines for RGB-D object recognition by using ResNet and compared them with our proposed approach. Motivated by the work of Gupta \textit{et al.}~\cite{DBLP:conf/eccv/GuptaGAM14} and Eitel \textit{et al.}~\cite{DBLP:conf/iros/EitelSSRB15}, we encoded each depth image into surface normal to efficiently exploit the information provided by depth image.

The structure of our different baseline methods is detailed as follows: 1) ResNet using only RGB image as input with and without pre-train model; 2) ResNet using only depth images as input; 3) ResNet using only surface normal images as input with and without pre-train model; and 4) two separate ways of ResNet trained for RGB and surface normal with last pooling layer concatenated, with and without pre-train model.

Table~1 shows the performance of different baselines. We see that the ResNet trained by raw depth images achieves worse performance than that trained by the surface normal. This is because surface normal can better represent geometry information than depth images. The accuracy rises swiftly when we combine the RGB-ResNet and SN-ResNet into a two way architecture.The experimental results clearly shows that the two way ResNet structure prone to be more effective and accurate with more information from modal-specific part.

In order to boost performance on the RGB-D object dataset, we also used the caffe model~\cite{DBLP:journals/corr/HeZRS15} pre-trained on the ImageNet~\cite{deng2009imagenet} for both the RGB input ResNet and the SN input ResNet. In contrast with the same structure without pre-train model, we achieved higher recognition accuracy.

\begin{table}[tb]
\centering
\caption{Comparison of our method and deep learning baselines.} \vspace*{6pt}
\begin{tabular}{|c|c|l|} \hline
Method& Accuracy (\%)\\ \hline
RGB ResNet & $81.6 \pm 1.9$\\ \hline
Depth ResNet & $79.2 \pm 1.7$\\ \hline
SN ResNet & $81.3 \pm 1.7$\\ \hline
RGB ResNet (pretrain) & $87.3 \pm 1.6$ \\ \hline
SN ResNet (pretrain) & $84.2 \pm 1.7$ \\ \hline
RGB-SN ResNet (two-way)  & $87.0 \pm 1.8$ \\ \hline
RGB-SN ResNet (two-way, pretrain)  & $90.5 \pm 1.7$ \\ \hline
Ours & $\bm{92.4 \pm 1.8}$ \\ \hline
\end{tabular}
\end{table}

\begin{table}[tb]
\centering
\caption{Comparison of our method and state-of-the-art methods on the RGB-D object dataset.}  \vspace*{6pt}
\begin{tabular}{|c|c|l|} \hline
Method& Accuracy (\%)\\ \hline
Lai~\textit{et al.}~\cite{lai_icra11a} & $81.9 \pm 2.8$\\ \hline
Blum~\textit{et al.}~\cite{blumicra2012} & $86.4 \pm 2.3$\\ \hline
Socher~\textit{et al.}~\cite{SocherEtAl2012:CRNN} & $86.8 \pm 3.3$\\ \hline
Bo~\textit{et al.}~\cite{Bo12unsupervisedfeature} & $87.5 \pm 2.9$\\ \hline
Wang~\textit{et al.}~\cite{Wang_2015_ICCV} & $88.5 \pm 2.2$\\ \hline
Eitel~\textit{et al.}~\cite{DBLP:conf/iros/EitelSSRB15} & $91.3 \pm 1.4$\\ \hline
Ours & $\bm{92.4 \pm 1.8}$ \\ \hline
\end{tabular}
\end{table}

\begin{figure}[tb]
        \centering
        \includegraphics[width = 0.4\textwidth]{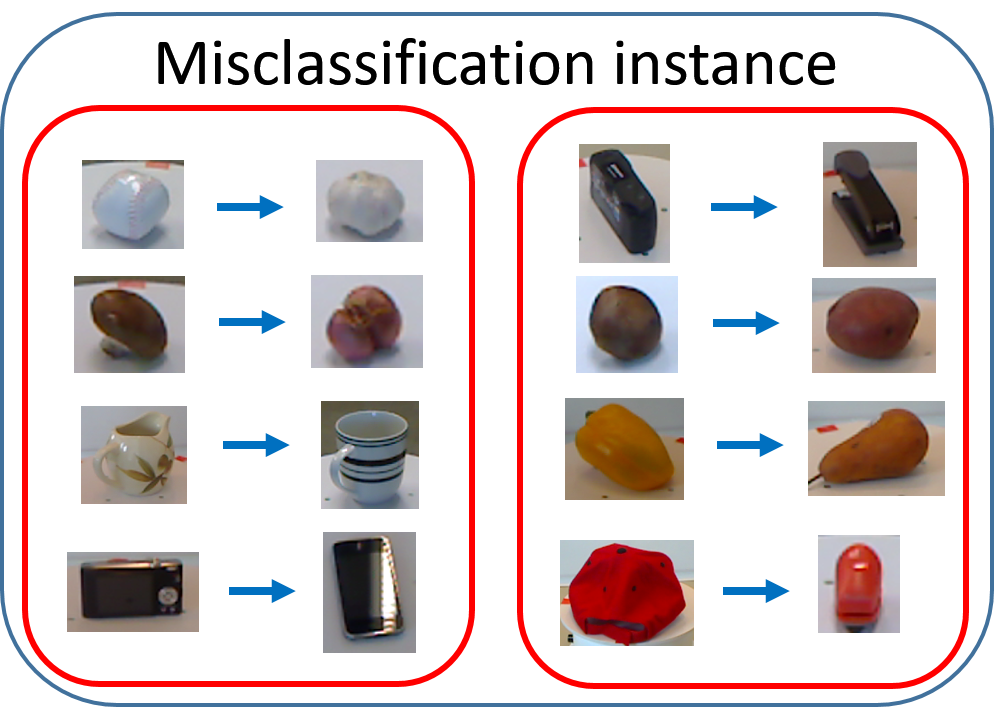}
        \caption{Some misclassification instances of our method in RGB-D object Dataset. Those misclassifications are caused by the large variations of shape, color or texture affinity between objects from other classes.}
\end{figure}

%
In our method, we used the surface normal as a substitute of the depth image with pre-trained model. Our proposed multi-modal learning method outperforms the best baseline by $1.9\%$ in terms of the accuracy. Comparing with the best baseline that simply connect the last pooling layer, our method can 1) generate the correlated part and the modal-specific part of different modalities and make sure they are not contaminated by each other, and 2) better exploit the information from the correlated and individual part of different modalities.

\textbf{Comparison with State-of-the-arts:} We also compare our method with six state-of-the-art RGB-D object recognition methods: 1) extracting depth feature using SIFT and spin images, RGB image using SIFT and color and texton histogram~\cite{lai_icra11a}; 2) CKM~\cite{blumicra2012}; 3) CNN-RNN~\cite{SocherEtAl2012:CRNN};4) HMP~\cite{Bo12unsupervisedfeature} 5) MMSS~\cite{Wang_2015_ICCV} and 6) FusNet~\cite{DBLP:conf/iros/EitelSSRB15}. Table~2 shows that our method outperformed all the state-of-the-art methods. We clearly see that our methods outperforms existing state-of-the-art methods.

Figure~3 shows the confusion matrix of recognition results on RGB-D object dataset. The diagonal elements represent the accuracy for each object class. We display several misclassified objects in Figure~4. Those error dues to the similarity objects from different classes like camera and cellphone or color affinity of ball and garlic. Texture similarity is also to blamed for misclassification.

\subsection{Results on 2D3D Dataset}

We utilized the same architecture in section 4.3 and used the same pre-trained caffe model. Table~3 shows the performance of different baselines on 2D3D dataset. Table~4 shows the comparison between our proposed method and several state-of-the-art methods. The confusion matrix of our recognition results is shown in Figure~5. Both among those baseline methods and other previous methods, we achieved better results.
\begin{figure}[tb]
        \centering
        \includegraphics[width = 0.4\textwidth]{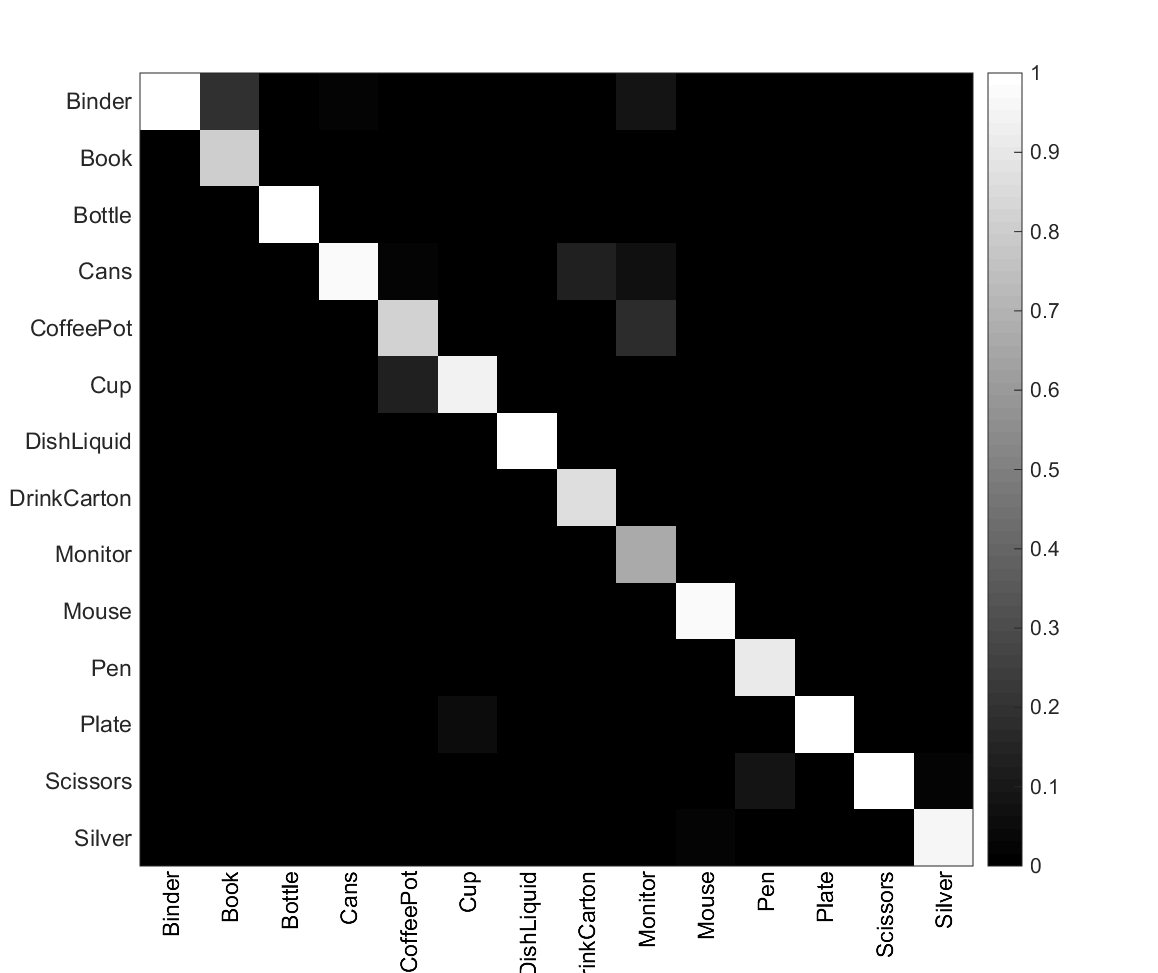}
        \caption{The confusion matrix of our method on the 2D3D dataset, where the vertical axis indicates the ground truth and the horizontal axis indicates the predicted labels, respectively.}\end{figure}

\begin{table}[tb]
\centering
\caption{Comparison with Existing Deep Learning Baselines on the 2D3D dataset.}  \vspace*{6pt}
\begin{tabular}{|c|c|l|} \hline
Method&Accuracy(\%)\\ \hline
RGB ResNet & $82.5$\\ \hline
SN ResNet & $81.6$\\ \hline
RGB ResNet(pretrain) & $86.4$\\ \hline
SN ResNet(pretrain) & $88.1$\\ \hline
RGB+SN ResNet (two-way)& $87.9$ \\ \hline
RGB+SN ResNet(two-way, pretrain) & $90.9$ \\ \hline
ours & $\bm{93.8}$ \\ \hline
\end{tabular}
\end{table}

\begin{table}[tb]
\centering
\caption{Comparison with state-of-the-art methods on the 2D3D dataset.}  \vspace*{6pt}
\begin{tabular}{|c|c|l|} \hline
Method&Accuracy(\%)\\ \hline
Browatzki~\textit{et al.}~\cite{conf/iccvw/BrowatzkiFGBW11} & $82.8$\\ \hline
Bo~\textit{et al.}~\cite{Bo12unsupervisedfeature} & $91.0$\\ \hline
Wang~\textit{et al.}~\cite{Wang_2015_ICCV} & $91.3$\\ \hline
Ours & $\bm{93.8}$\\ \hline
\end{tabular}
\end{table}

\subsection{Parameter Analysis}

The accuracy of our object recognition method is affected by the weights $c_i(i=1,2,3)$, which decides whether the correlated component or the modal-specific part from RGB or depth dominates. In our proposed method, the weights are self-adapted. In this section, we kept the weights $c_i(i=1,2,3)$ fixed and revealed the relationship between weights and the recognition accuracy. Note that $c_1$ is for the correlated part and $c_2, c_3$ correspond to the RGB and depth modality, respectively, where we have $\sum_{i=1} ^{3}{c_i}=1,c_i>0$. We set parameter $p$ as $p = c_1$.

Figure~6 shows the recognition rate of our method versus different values of the weighting parameter $p$. In Figure~6, when $p$ is small, which means that the correlated part plays smaller significance in recognition and the accuracy is relatively low. When $p$ becomes larger, the correlated part gradually plays a relatively more important part than $p$ is low, the accuracy rises. However, when $p$ is too large, which means that the correlated part dominates, the accuracy decreases as the effects of the modal-specific part begin to vanish. The value of $c$ reaches the peak is close to the scenario when the parameter $c$ learned by our multi-modal learning method.

\begin{figure}[tb]
        \centering
        \includegraphics[width = 0.4\textwidth]{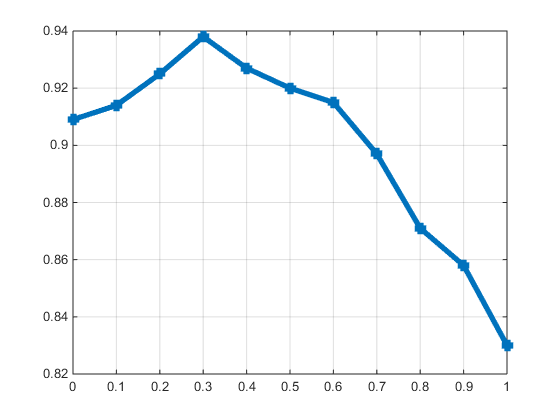}
        \caption{The performance relationship between $p$ and the recognition accuracy on the RGB-D object dataset.}
\end{figure}

\section{Conclusions}

In this paper, we have proposed a correlated and individual multi-modal deep learning method for RGB-D object recognition. In our proposed method, we enforced both the correlated and the modal-specific parts in our learned features for RGB-D image object to satisfy several characteristics within a joint learning framework, so that the sharable and modal-specific information can be well exploited. Experimental results on two widely used RGB-D object image benchmark datasets clearly show that our method outperforms state-of-the-art methods.

{\small
\bibliographystyle{ieee}
\bibliography{egbib}
}

\end{document}